\documentclass[compsoc,conference,a4paper,10pt,times]{IEEEtran}
\IEEEoverridecommandlockouts
\usepackage{cite}
\usepackage{amsmath,amssymb,amsfonts}
\usepackage{algorithmic}
\usepackage{graphicx}
\usepackage{MnSymbol}
\usepackage{tikz}
\usetikzlibrary{matrix}
\usetikzlibrary{positioning}
\usetikzlibrary{shapes}
\usepackage{textcomp}
\usepackage{bmpsize}
\usepackage{xcolor}
\usepackage{lipsum}
\usepackage{comment}
\usepackage[colorlinks=true,urlcolor=black]{hyperref}
\def\BibTeX{{\rm B\kern-.05em{\sc i\kern-.025em b}\kern-.08em
    T\kern-.1667em\lower.7ex\hbox{E}\kern-.125emX}}
\begin{document}

\title{Distribution and Clusters Approximations as Abstract Domains in Probabilistic Abstract Interpretation to Neural Network Analysis\\ {\footnotesize
    \textsuperscript{}} \thanks{}  }

\author{\IEEEauthorblockN{1\textsuperscript{st} {
Zhuofan Zhang}}
\IEEEauthorblockA{\textit{Department of Computing} \\
\textit{Imperial College London}\\
London, UK \\
zhuofan.zhang13@imperial.ac.uk}
\and
\IEEEauthorblockN{2\textsuperscript{nd} Herbert Wiklicky}
\IEEEauthorblockA{\textit{Department of Computing} \\
\textit{Imperial College London}\\
London, UK \\
h.wiklicky@imperial.ac.uk}
}

\maketitle

\begin{abstract}
The probabilistic abstract interpretation framework of neural network analysis analyzes a neural network by analyzing its density distribution flow of all possible inputs. The grids approximation is one of abstract domains the framework uses which abstracts concrete space into grids. In this paper, we introduce two novel approximation methods: distribution approximation and clusters approximation. We show how these two methods work in theory with corresponding abstract transformers with help of illustrations of some simple examples.
\end{abstract}


\section{Introduction}
Different methods in the field of neural network verification and analysis are desired to help humans better understand how neural networks make their decisions and to ensure certain properties of neural networks, such as safety or robustness. One branch of this research has focused on using techniques of abstract interpretation theory. In 2018 Gehr et al. proposed an analyzer $AI^{2}$ that uses abstract interpretation to over-approximate a range of concrete values into an abstract domain and checks whether this abstract domain holds the expected result and which will prove every single input over-approximated by this abstract domain also holds this result \cite{ai2}. This analyzer makes use of the zonotope abstract domain which was originally proposed by Ghorbal et al. \cite{ghorbal}. Later, Singh et al. \cite{ai2_2} proposed a new abstract domain which combines floating point polyhedra with intervals, instead of the former zonotopes abstract domain. Its performance was compared with Reluplex, which is a former property verification technique that extends the simplex algorithm to support ReLU constraints \cite{reluplex}. Singh et al. also presented DeepZ based on abstract interpretation that handles ReLU, Tanh and Sigmoid activation functions and supports feedforward, convolutional, and residual architectures \cite{deepz}. Abstract interpretation is also bridged with gradient-based optimization and can be applied to training neural networks \cite{ai2ontrain}. Ryou et al. presented Prover, a scalable and precise verifier for recurrent neural networks using polyhedral abstractions \cite{prover}.\\
Beyond classic verifiers, the question of whether certain neural network properties hold in probabilistic conditions is also considered. Techniques of the theory of probabilistic abstract interpretation have been then brought into the research. Probabilistic abstract interpretation is a framework that allows to systematically lift any classical analysis or verification method to the probabilistic setting by separating in the program semantics the probabilistic behavior from the (non-)deterministic behavior \cite{paicousot}. Di Pierro and Wiklicky also developed the framework of probabilistic abstract interpretation by using Concurrent Constraint Programming as a reference programming paradigm in probabilistic program analysis \cite{ccp}, and by investigating relations between the operational semantics of probabilistic programming languages and Discrete Time Markov Chains (DTMCs) to introduce Linear Operator semantics(LOS) \cite{PAI}, and used the framework to compare analysis on the basis of their expected exactness for a given program with a quantitative notion of precision \cite{precision}, and also extended the framework from finite state spaces to infinite state spaces \cite{wla}. In 2020 Steffen et al. introduced NetDice which is a scalable and accurate probabilistic network configuration analyzer that is based on an inference algorithm to efficiently explore the space of failure scenarios \cite{netdice}. Mangal et al. verifies probabilistic robustness of neural networks  with respect to the input distribution \cite{mangal}. Pasareanu et al. investigated the use of symbolic analysis and constraint solution space quantification to precisely quantify probabilistic properties in neural networks \cite{pasareanu}. In 2024 Zhang and Wiklicky proposed a method of grids approximation to analyze neural networks with techniques from probabilistic abstract interpretation \cite{zhang}. \\
In this paper, we introduce two novel approximation methods: distribution approximation and clusters approximation based on the probabilistic abstract interpretation framework with respect to neural network analysis. Different approximation approaches may be chosen with respect to different program circumstances and different properties concerned. Our approach aims to analyze the probability density distribution flow of all possible inputs of a trained neural network. This can be used to make sure that the system is robust and under controlled in context of probability, by showing that a majority of data flows through propagation of neural networks are bounded in desired domain. In addition, it can also offer a sensitivity analysis of the data from the input to each layer, helping to better understand how the data flow inside a neural network. We show how these two novel approximation methods work in theory and illustrate the corresponding abstract transformers used in the framework with help of some examples.

\section{Abstract Domains}

\subsection{Distribution Approximation as Abstract Domain}
The first new approximation method we present is the distribution approximation. The distribution approximation approximates the concrete space in its distribution that includes all concrete data plus the corresponding probabilities. That is, if a concrete data point is in dimension $n$, then the approximated model is $n+1$ dimensional, the additional dimension being the probabilities of the concrete data. An abstract transformer in this case maps a distribution following propagations of neural network layers. In other words, we consider a distribution that propagates through the neural network in the abstract domain. In distribution approximation, different choices of function approximators can be selected with respect to problem circumstance. Polynomials, radial basis functions, or Fourier transform can be one of typical choices.
\subsubsection{Polynomial Regression}
Polynomials can be one of the simplest approximators in such a regression problem. A general polynomial model is in the form of 
\begin{equation}
    \textbf{y} = \beta_0 + \beta_1 \textbf{x} + \beta_2 \textbf{x}^2 + ... +  \beta_m \textbf{x}^m + \epsilon
\end{equation}
Or it can be written as a system of linear equations:
\begin{equation}
    \begin{bmatrix}
        y_1\\
        y_2\\
        \vdots \\
        y_n
    \end{bmatrix} = 
    \begin{bmatrix}
        1 & x_1 & x_1^2 & ... & x_1^m\\
        1 & x_2 & x_2^2 & ... & x_2^m\\
        \vdots & \vdots & \vdots & \ddots & \vdots\\
        1 & x_n & x_n^2 & ... & x_n^m\\
    \end{bmatrix}
    \begin{bmatrix}
        \beta_0\\
        \beta_1\\
        \beta_2\\
        \vdots\\
        \beta_m
    \end{bmatrix}
    + \begin{bmatrix}
        \epsilon_1\\
        \epsilon_2\\
        \vdots\\
        \epsilon_n
    \end{bmatrix}
\end{equation}
which can also be written in matrix notation of
\begin{equation}
    \overrightarrow{y} = \textbf{X}\overrightarrow{\beta} + \overrightarrow{\epsilon}
\end{equation}
and we can find the least square solution
\begin{equation}
    \hat{\overrightarrow{\beta}} = (\textbf{X}^{\intercal}\textbf{X})^{-1}\textbf{X}^{\intercal}\overrightarrow{y}
\end{equation}
We can now have the polynomials that approximate the concrete space. By definition of probabilistic abstract interpretation, $f$ in concrete space is an affine function representing the forward propagation of the neural network layer, therefore we can have the abstract transformer $f^{\#}$ in abstract domain mapping a distribution to a distribution such that
\begin{equation}
    f^{\#}(\textbf{y}(\textbf{x})) = \textbf{y}(f^{-1}(\textbf{x}))
\end{equation}

\subsubsection{Radial Basis Functions}
Radial basis functions(RBFs) are real-valued functions whose value depends only on the distance from a central point, typically expressed as
\begin{equation}
    \phi(\textbf{x}) = \phi(\|\textbf{x}-\textbf{c}\|)
\end{equation}
where \textbf{x} is the input vector, \textbf{c} is the center, and $\|\cdot\|$ denotes a norm(often Euclidean). Common RBFs include the Gaussian function $\phi(\textbf{x}, \textbf{c}) = e^{-\frac{\|\textbf{x} - \textbf{c}\|^2}{2\sigma^2}}$ where $\sigma$ is a parameter that controls the width of the basis function, multiquadric function $\phi(\textbf{x}, \textbf{c}) = \sqrt{\|\textbf{x} - \textbf{c}\|^2 + \sigma^2}$, inverse multiquadric function $\phi(\textbf{x}, \textbf{c}) = \frac{1}{\sqrt{\|\textbf{x} - \textbf{c}\|^2 + \sigma^2}}$, or thin plate spline function $\phi(\textbf{x}, \textbf{c}) = \|\textbf{x} - \textbf{c}\|^2 \ln(\|\textbf{x} - \textbf{c}\|)$.\\
More specifically, radial basis functions approximate the concrete space in a distribution with $N$ centers. That is, if let original concrete space be $\textbf{x}$, and let the dimension of probabilities be $y$, the approximation can be represented in the form of
\begin{equation}
    \textbf{y} = \sum_{i = 1}^{N} a_i\phi(\|\textbf{x}-\textbf{c}\|)
\end{equation}
where $a_i$ are the interpolation coefficients of the radial basis functions. Note that $a_i$ can be computed once given a particular choice of $N$ centers. To find the optimal choice of centers, we want the loss function
\begin{equation}
    RMS = \sqrt{\frac{1}{N_{data}}\sum_{i=1}^{N_{data}}(y_{model_{i}} - y_{data_{i}})^2}
\end{equation}
to be minimized \cite{rbf}.\\
Once we have computed the radial basis functions that approximate the concrete space, we can have the abstract transformer $f^{\#}$ in the abstract domain mapping a distribution to a distribution such that
\begin{equation}
    f^{\#}(\textbf{y}(\textbf{x})) = \textbf{y}(f^{-1}(\textbf{x}))
\end{equation}
\textbf{Example } Consider a one-dimensional example containing 13 concrete data points being one-dimensional values $\{0.1, 0.4, 0.5, 0.8, 1.5, 2.1, 3.0, 3.1, 3.5, 4.6, 5.9, 6.0, 6.4\}$, and together with their probabilities $\{0.072, 0.076, 0.08, 0.073, 0.036, 0.014, 0.02, 0.012, $\\$0.016, 0.02, 0.022, 0.03, 0.024\}$ respectively. We say that we wish to approximate these data points by radial basis functions with three centers. That is, we approximate these data by
\begin{equation}
    y = \sum_{i = 1}^{3} a_i\phi(\|\textbf{x}-\textbf{c}\|)
\end{equation}
Also, let the radial basis functions be Gaussian functions
\begin{equation}
    \phi(\|\textbf{x}-\textbf{c}\|) = e^{-\|\textbf{x}-\textbf{c}\|^2}
\end{equation}
We first have to decide which three centers to choose among the 13 data points. In an exhaustive search, models must be constructed for every possible combination of centers. In each case, we need to solve the system of linear equations
\begin{equation}
    \sum_{i = 1}^{N} a_i\phi(\|\textbf{x}_k-\textbf{c}_i\|) = y_k
\end{equation}
That is, in this case
\begin{equation}
    \begin{pmatrix}
        \phi(\|\textbf{x}_1-\textbf{c}_1\| & \phi(\|\textbf{x}_2-\textbf{c}_1\| & \phi(\|\textbf{x}_3-\textbf{c}_1\|\\
        \phi(\|\textbf{x}_1-\textbf{c}_2\| & \phi(\|\textbf{x}_2-\textbf{c}_2\| & \phi(\|\textbf{x}_3-\textbf{c}_2\| \\
        \phi(\|\textbf{x}_1-\textbf{c}_3\| & \phi(\|\textbf{x}_2-\textbf{c}_3\| & \phi(\|\textbf{x}_3-\textbf{c}_3\|
    \end{pmatrix} 
    \begin{pmatrix}
        a_1 \\
        a_2 \\
        a_3 
    \end{pmatrix} =
    \begin{pmatrix}
        y_1\\
        y_2\\
        y_3
    \end{pmatrix}
\end{equation}
to get interpolation coefficients $a_i$. Then the cost function are evaluated over the full dataset, and the optimal model is selected with the smallest value of the cost function. The total number of combinations in this case is
\begin{equation}
    \begin{pmatrix}
        13\\
        3
    \end{pmatrix} = 
    \frac{13!}{3!(13-3)!} = 286
\end{equation}
Therefore, we exhaustively search for 286 choices of combinations of centers and for each choice we compute loss function $RMS$ which has been introduced above. We finally find that the smallest value of cost function is achieved when the selected combination of centers are $(0.5, 0.08), (3.0, 0.02), (6.0, 0.03)$, and the corresponding RBFs is
\begin{equation}
    y = 0.8 * e^{-(x-0.5)^2} + 0.2 * e^{-(x-3)^2} + 0.3 * e^{-(x-6)^2}
\end{equation}
Now say we have a very simple neural network layer with only one input node and one output node such that
\begin{equation}
    f(x) = 2x + 1 
\end{equation}
In abstract domain of probabilistic abstract interpretation framework we have 
\begin{equation}
    f^{\#}(X^{\#}) = f^{\#}(A(X)) = f^{\#}(y(X)) = y(f^{-1}(X))
\end{equation}
by formula of abstract transformer of the distribution approximation given in (9). Then we can see that
\begin{equation}
    \begin{split}
        y(f^{-1}(X)) &= 0.8 * e^{-(\frac{x-1}{2}-0.5)^2} + 0.2 * e^{-(\frac{x-1}{2}-3)^2}\\
        &+ 0.3 * e^{-(\frac{x-1}{2}-6)^2}
    \end{split}
\end{equation}
which is the distribution approximation after propagation of the layer.\\
\\
Using radial basis functions as an abstract domain in probabilistic abstract interpretation has several advantages. Firstly, radial basis functions is a powerful approximator which can well approximate any continuous function given appropriate centers and parameters. Secondly, 
radial basis functions can capture the behavior of the distribution with precise localized effects, which clearly shows the local features of the distribution. Also, radial basis functions are a smooth approximation and can fit well for composite affine functions.

\subsubsection{Fourier Transform}
The Fourier transform is another practical choice of an approximator for distribution approximation. It is a mathematical operation that transforms a function of time (or space) into a function of frequency. A Fourier transform $F(\omega)$ of its original function $f(t)$ is normally expressed as \cite{fourier}
\begin{equation}
    F(\omega) = \int_{-\infty}^{\infty} f(t) e^{- j\omega t} dt
\end{equation}
Or we can write it as the inverse Fourier transform
\begin{equation}
    f(t) = \int_{-\infty}^{\infty} F(\omega) e^{j\omega t} d\omega
\end{equation}
For a probability distribution, the Fourier transform provides a characteristic function which encodes the distribution's properties in the domain of frequency. That is, it captures and grabs information from all different frequencies contained in this original distribution. This is the same purpose for which the abstraction operation in probabilistic abstract interpretation framework is seeking.\\
Note that since the Fourier transform captures features of frequency of a continuous function, if the concrete space is not in such continuous form, it is necessary to take abstraction of concrete space into a continuous distribution first, for example by the radial basis function approximation, and then followed by the Fourier transform approximation. That is, a two-step abstraction is taken in this probabilistic abstract interpretation, and only the second abstract domain is considered. 
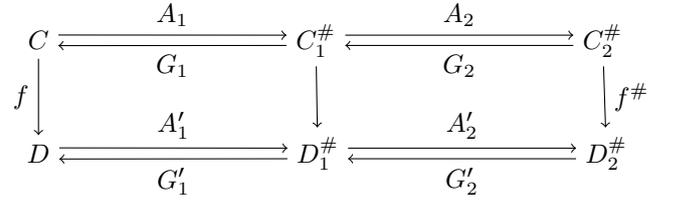
\begin{figure}[h]
\centering
\begin{tikzpicture}[node distance=1.5cm]
\node(C)                            {$C$};
\node(D)     [below of=C]           {$D$};
\node(Csharp)     [right =3cm of C]     {$C_1^{\#}$};
\node(Dsharp)     [right =3cm of D]     {$D_1^{\#}$};
\node(Csharp2)     [right =3cm of Csharp]     {$C_2^{\#}$};
\node(Dsharp2)     [right =3cm of Dsharp]     {$D_2^{\#}$};

\draw[->](C)         -- node[left]{$f$} (D);
\draw[->](Csharp)         -- node[right]{$ $} (Dsharp);
\draw[->](C.15)         -- node[above]{$A_1$}(Csharp.170);
\draw[<-](C.345)         -- node[below]{$G_1$}(Csharp.190);
\draw[->](D.15)         -- node[above]{$A_1'$}(Dsharp.170);
\draw[<-](D.345)         -- node[below]{$G_1'$}(Dsharp.190);

\draw[->](Csharp2)         -- node[right]{$f^{\#}$} (Dsharp2);
\draw[->](Csharp.10)         -- node[above]{$A_2$}(Csharp2.170);
\draw[<-](Csharp.350)         -- node[below]{$G_2$}(Csharp2.190);
\draw[->](Dsharp.10)         -- node[above]{$A_2'$}(Dsharp2.170);
\draw[<-](Dsharp.350)         -- node[below]{$G_2'$}(Dsharp2.190);

\end{tikzpicture}
\caption{probabilistic abstract interpretation with two-step abstraction}
\end{figure}

Once we have obtained the Fourier transform that captures features of the distribution that approximate the concrete space, we can have the abstract transformer $f^{\#}$ in the abstract domain of frequency with respect to the propagation of the neural network layer such that
\begin{equation}
    f^{\#}(F(\omega)) = F(g(\omega))
\end{equation}
where $g(\omega)$ is a function of the frequency change deduced from the affine function $f$. \\
\\
\textbf{Example } Take the same example illustrated in radial basis functions approximation section that the distribution approximating concrete space is given
\begin{equation}
    y = 0.8 * e^{-(x-0.5)^2} + 0.2 * e^{-(x-3)^2} + 0.3 * e^{-(x-6)^2}
\end{equation}
and the one-input node neural network layer is 
\begin{equation}
    f(x) = 2x + 1 
\end{equation}
We first take the Fourier transform of function $y(x)$ such that
\begin{equation}
    \begin{split}
        &F(\omega) = \int_{-\infty}^{\infty} y(x) e^{- j\omega x} dx\\
        &= \int_{-\infty}^{\infty} (0.8 * e^{-(x-0.5)^2} + 0.2 * e^{-(x-3)^2} + 0.3 * e^{-(x-6)^2}) e^{- j\omega x} dx\\
        & = \int_{-\infty}^{\infty} 0.8 * e^{-(x-0.5)^2}e^{- j\omega x} dx + \int_{-\infty}^{\infty} 0.2 * e^{-(x-3)^2}e^{- j\omega x} dx \\
        &+ \int_{-\infty}^{\infty} 0.3 * e^{-(x-6)^2}e^{- j\omega x} dx
    \end{split}
\end{equation}
let $t_1 = x - 0.5$, $t_2 = x - 3$, $t_3 = x - 6$,
\begin{equation}
    \begin{split}
    F(\omega) &= \int_{-\infty}^{\infty} 0.8 * e^{-t_1^2}e^{- j\omega (t_1 + 0.5)} dt_1 + \int_{-\infty}^{\infty} 0.2 * e^{-t_2^2}e^{- j\omega (t_2+3)} dt_2 \\
    &+ \int_{-\infty}^{\infty} 0.3 * e^{-t_3^2}e^{- j\omega (t_3+6)} dt_3
    \end{split}
\end{equation}
Consider only the first term for convenience, and the other two are similar
\begin{equation}
    \begin{split} 
        F_1(\omega) &= \int_{-\infty}^{\infty} 0.8 * e^{-t_1^2}e^{- j\omega (t_1 + 0.5)} dt_1\\
        &= 0.8*\int_{-\infty}^{\infty} e^{-t_1^2}cos(\omega (t_1+0.5)) dt_1 \\
        &+ 0.8 * j \underbrace{\int_{-\infty}^{\infty} e^{-t_1^2}sin(\omega (t_1+0.5)) dt_1}_{0} \\
        &= 0.8*\int_{-\infty}^{\infty} e^{-t_1^2}cos(\omega (t_1+0.5)) dt_1 \\
        &= 0.8\sqrt{\pi} e^{-\omega /4}
    \end{split}
\end{equation}
where detailed deduction can be found in \cite{fourier2}. We can see that the Fourier transform of a Gaussian distribution is still a Gaussian distribution. That is, this abstraction does not capture additional information, therefore in this case the Fourier transform may not be an appropriate approximator.\\
\\
Using the Fourier transform as a distribution approximator in probabilistic abstract interpretation is especially powerful for capturing features of frequency including amplitude, phase and harmonics which are three key components. It is also a universal approximator for the distribution function as any function can be regarded as a periodic function by assuming that the period is infinitely long. However, it is still more appropriate for approximating periodic functions. In addition, it is normally used as a second-step abstraction, given that the concrete space has already been approximated to a continuous function in the first-step abstraction.

\subsection{Clusters Approximation as Abstract Domain}
Clusters approximation is an approximation method with abstract domains where all possible concrete input data points are clustered into a number of groups, with only the centroid of each group being abstract element representing each group of concrete data points. The grids approximation \cite{zhang} can also be regarded as a special kind of clusters approximations. Considering clustering is a typical unsupervised machine learning problem, there are some commonly used unsupervised machine learning methods that also can be used in this case. 
\subsubsection{K-means}
K-means clustering is an unsupervised machine learning algorithm used for partitioning a dataset into a pre-defined number of groups. It operates by categorizing data points into clusters by using a mathematical distance measure, usually euclidean, from the cluster center. The objective is to minimize the sum of distances between data points and their assigned clusters. Data points that are nearest to a centroid are grouped together within the same category. \\
To train and obtain a K-means clustering model, the first step is to initialize $K$ centroids, where $K$ is equal to the number of clusters chosen for a specific dataset. Then it is computed by a two-step iterative process based on the expectation-maximization machine learning algorithm. The expectation step assigns each data point to its closest centroid according to distance. The maximization step computes the mean of all the points for each cluster and reassigns the cluster center, or centroid. This process repeats until the centroid positions have reached convergence or the maximum number of iterations has been reached.\\
Given the dataset $x^{(1)}, ..., x^{(m)}$ and the labels of data points $c^{(1)}, ..., c^{(m)}$, the algorithm is as follows:\\
1. Initialize cluster centroids $\mu_1, \mu_2, ..., \mu_K \in \mathbb{R}^n$ at random.\\
2. Repeat until convergence: \\
    \begin{equation}
        \text{For every } i \text{, set } c^{(i)} := \arg\min_{k} \|x^{(i)} - \mu_k\|^2.
    \end{equation}
    \begin{equation}
        \text{For each } k \text{, set } \mu_k := \frac{\sum^m_{i=1}1\{c^{(i)}==k\}x^{(i)}}{\sum^m_{i=1}1\{c^{(i)}==k\}}.
    \end{equation}
Eventually, the iteration converges and we get $K$ centroids of all $K$ clusters. These $K$ clusters $c^{\#}_1, ..., c^{\#}_K \in C^{\#}$ form the abstract domain $C^{\#}$. That is, there is an abstraction function $A$ such that
\begin{equation}
    A(\{x^{(i)} \mid c^{(i)}==k\}) = c^{\#}_k
\end{equation}
for each $k = 1, ..., K$. Meanwhile, if given the probabilities that the data points $x^{(1)}, ..., x^{(m)}$ are $p_{x^{(1)}}, ..., p_{x^{(m)}}$, we have 
\begin{equation}
    p_{c^{\#}_k} = \sum_i p_{x^{(i)}}\{c^{(i)}==k\}
\end{equation}
and the abstract transformer performing on an abstract element is in fact the concrete function performing on the cluster centroid $f^{\#}: C^{\#} \rightarrow D^{\#}$
\begin{equation}
    f^{\#}(c^{\#}_k) = f(\mu_k)
\end{equation}
for each $k = 1, ..., K$.\\
\\
\textbf{Example } Let's take the previously illustrated example. Given function $f : \mathbb{R}^2 \rightarrow \mathbb{R}^2$
\begin{equation}
    f(x) = \begin{pmatrix}
        2 & -1\\
        0 & 1
    \end{pmatrix} x
\end{equation}
with input space X expressed by the zonotope $z^1 :[-1, 1]^3 \rightarrow \mathbb{R}^2$
\begin{equation}
    z^1(\epsilon_1, \epsilon_2, \epsilon_3) = (1 + 0.5\epsilon_1 + 0.5\epsilon_2, 2+0.5\epsilon_1 + 0.5\epsilon_3)
\end{equation}
The geometric expression of the input space can be found in Figure 2. In this input space, there are infinitely many data points equally distributed. Let us choose number of clusters in this case to be two, and first randomly initialize two centroids of clusters, say $(0,1)$ and $(2,3)$. Then the iteration step follows until it converges to find two centroids of clusters. To simplify the process, let us say after a number of iterations two centroids eventually converge to $(\frac{11}{18}, \frac{29}{18})$ and $(\frac{25}{18}, \frac{43}{18})$. 
\begin{figure}[h]
\centering
\begin{tikzpicture}[scale=0.8]
\draw[fill=gray!22] (0,1) -- (1,1) -- (2,2) -- (2,3) -- (1,3) -- (0,2) -- (0,1);

\draw[->] (-2,0) -- (3,0) coordinate (x axis);
\draw[->] (0,-1) -- (0,4) coordinate (y axis);

\draw (0,1) -- (0,1) node[anchor=east] {$1$};
\draw (0,2) -- (0,2) node[anchor=east] {$2$};
\draw (0,3) -- (0,3) node[anchor=east] {$3$};

\draw (-1,0) -- (-1,0) node[anchor=north] {$-1$};
\draw (1,0) -- (1,0) node[anchor=north] {$1$};
\draw (2,0) -- (2,0) node[anchor=north] {$2$};

\draw[blue] (11/18, 29/18) circle (2pt);
\draw[blue] (25/18, 43/18) circle (2pt);

\end{tikzpicture}
\caption{Two centroids of clusters}
\end{figure}
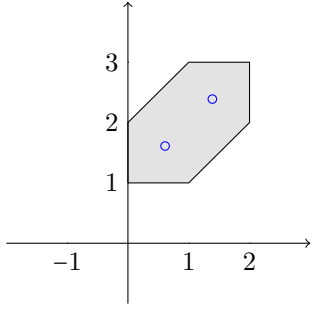
That is, these two centroids as abstract elements take an approximation of the input space by
\begin{equation}
    \begin{split}
    A(\{x^{(i)} \mid c^{(i)}==1\}) = c^{\#}_1 = (\frac{11}{18}, \frac{29}{18})\\
    \quad A(\{x^{(i)} \mid c^{(i)}==2\}) = c^{\#}_2 = (\frac{25}{18}, \frac{43}{18})
    \end{split}
\end{equation}
To compute layer propagation in the abstract domain, we have the abstract transformer performing on two centroids such that $f^{\#}: C^{\#} \rightarrow D^{\#}$
\begin{equation}
    f^{\#}(c^{\#}_1) = f(\mu_1) = \begin{pmatrix}
        2 & -1\\
        0 & 1
    \end{pmatrix} \begin{pmatrix}
        \frac{11}{18} \\
        \frac{29}{18} 
    \end{pmatrix} = (-\frac{7}{18}, \frac{29}{18})
\end{equation}
\begin{equation}
    f^{\#}(c^{\#}_2) = f(\mu_2) = \begin{pmatrix}
        2 & -1\\
        0 & 1
    \end{pmatrix} \begin{pmatrix}
        \frac{25}{18} \\
        \frac{43}{18} 
    \end{pmatrix} = (\frac{7}{18}, \frac{43}{18})
\end{equation}
\begin{figure}[h]
\centering
\begin{tikzpicture}[scale=0.8]
\draw[fill=gray!22] (1,1) -- (2,2) -- (1,3) -- (-1,3) -- (-2,2) -- (-1,1) -- (1,1);

\draw[->] (-3,0) -- (3,0) coordinate (x axis);
\draw[->] (0,-1) -- (0,4) coordinate (y axis);

\draw (0,1) -- (0,1) node[anchor=east] {$1$};
\draw (0,2) -- (0,2) node[anchor=east] {$2$};
\draw (0,3) -- (0,3) node[anchor=east] {$3$};

\draw (-2,0) -- (-2,0) node[anchor=north] {$-2$};
\draw (-1,0) -- (-1,0) node[anchor=north] {$-1$};
\draw (1,0) -- (1,0) node[anchor=north] {$1$};
\draw (2,0) -- (2,0) node[anchor=north] {$2$};

\draw[blue] (-7/18, 29/18) circle (2pt);
\draw[blue] (7/18, 43/18) circle (2pt);

\end{tikzpicture}
\caption{Two centroids of clusters after layer propagation}
\end{figure}
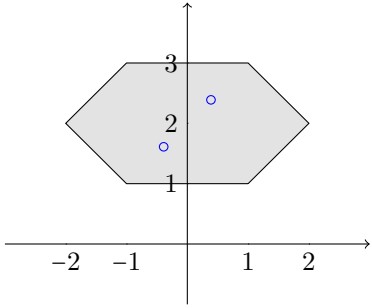
The corresponding probabilities of two centroids depend on the probabilities with which the input space is distributed. If the probability dimension is initialized with a uniform distribution, then two converged centroids straightforwardly will have
\begin{equation}
    p_{c^{\#}_1} = \sum_i p_{x^{(i)}}\{c^{(i)}==1\} = \frac{1}{2}, p_{c^{\#}_2} = \sum_i p_{x^{(i)}}\{c^{(i)}==2\} = \frac{1}{2}
\end{equation}
Note that after layer propagation the resulting coordinates of two centroids are not equal to intuitive centroids of resulting shapes geometrically. That is because the affine transformation changes the distances between the points and therefore changes the density of the shape, which is not straightforward in infinitely distributed case. The two centroids as abstract elements in clustering approximation in this case have better stored information from the previous layer.

\subsubsection{Gaussian Mixture Model}
The Gaussian mixture model(GMM) is another commonly used clustering method in unsupervised machine learning. A Gaussian mixture model is a parametric probability density function represented as a weighted sum of Gaussian component densities. Unlike K-means which is a `hard' clustering method which means that it will associate each point to one and only one cluster, the Gaussian mixture model is a soft clustering method used to determine the probabilities each data point belongs to several possible clusters. Similarly to K-means, parameters of a Gaussian mixture model are also estimated using iterative Expectation-Maximization algorithm from training data.\\
More specifically, a Gaussian mixture is a function that is composed of $K$ Gaussians, identified by $k \in {1, ..., K}$. Each Gaussian is comprised of a mean $\mu_k$ that defines its center, a covariance $\Sigma_k$ that defines its width, and a mixture weight $w_k$ that defines how big or small the Gaussian function will be. That is, a Gaussian mixture model is a weighted sum of K Gaussian densities as given by the equation \cite{gmm}
\begin{equation}
    p(\mathbf{x} | \lambda_k) = \sum_{k=1}^{K} w_k \mathcal{N}(\mathbf{x} \mid \mu_k, \Sigma_k)
\end{equation}
where $\lambda_k = \{w_k, \mu_k, \Sigma_k\}, k = 1, ..., K$ is the notation of all parameters and each Gaussian density is a Gaussian function of the form
\begin{equation}
    \mathcal{N}(\mathbf{x} \mid \mu_k, \Sigma_k) = \frac{1}{(2\pi)^{D/2}|\Sigma_k|^{1/2}} e^{-\frac{1}{2}(\mathbf{x}-\mu_k)^\text{T}\Sigma_k^{-1}(\mathbf{x} - \mu_k)}
\end{equation}
where $D$ is the number of dimensions of each data point. The mixture weights must satisfy the constraint such that
\begin{equation}
    \sum_{k=1}^{K} w_k  = 1
\end{equation}
Given the dataset $x^{(1)}, ..., x^{(m)}$, the iterative expectation-maximization algorithm is as follows:\\
1. E-step: for each data point $x^{(i)}$, compute $r_{ik}$ the probability that it belongs to cluster $k$
\begin{equation}
    r_{ik} = \frac{w_k \mathcal{N}(x^{(i)} \mid \mu_k, \Sigma_k)}{\sum_{k'} w_{k'} \mathcal{N}(x^{(i)} \mid \mu_{k'}, \Sigma_{k'})}
\end{equation}
2. M-step: for each cluster $k$, update its parameters using the $r_{ik}$ computed from previous step
\begin{equation}
    m_{k} = \sum_i r_{ik}, \quad w_k = \frac{m_k}{m}
\end{equation}
\begin{equation}
    \mu_k = \frac{1}{m_k} \sum_i r_{ik} x^{(i)}, \quad \Sigma_k = \frac{1}{m_k} \sum_i r_{ik} (x^{(i)} - \mu_k)^T (x^{(i)} - \mu_k)
\end{equation}
Eventually, the iteration converges and the convergence is guaranteed. Again, these $K$ Gaussians $c^{\#}_1, ..., c^{\#}_K \in C^{\#}$ form the abstract domain $C^{\#}$, with
\begin{equation}
    c^{\#}_k = <\mu_k, \Sigma_k, w_k>
\end{equation}
If given the probabilities of the data points $x^{(1)}, ..., x^{(m)}$ are $p_{x^{(1)}}, ..., p_{x^{(m)}}$, we have 
\begin{equation}
    p_{c^{\#}_k} = \sum_i r_{ik} p_{x^{(i)}}
\end{equation}
and the abstract transformer is
\begin{equation}
    \begin{split}
         f^{\#}(c^{\#}_k) &= <f^{\#}(\mu_k), f^{\#}(\Sigma_k), w_k> \\
         & = <A\mu_k + b, A\Sigma_k A^\textbf{T}, w_k> 
    \end{split}
\end{equation}
given that the neural network layer propagation is $f(\textbf{x}) = A\textbf{x} + b$. \\
\\
\textbf{Example } Again taking the function $f : \mathbb{R}^2 \rightarrow \mathbb{R}^2$
\begin{equation}
    f(x) = \begin{pmatrix}
        2 & -1\\
        0 & 1
    \end{pmatrix} x
\end{equation}
with input space X expressed by the zonotope $z^1 :[-1, 1]^3 \rightarrow \mathbb{R}^2$
\begin{equation}
    z^1(\epsilon_1, \epsilon_2, \epsilon_3) = (1 + 0.5\epsilon_1 + 0.5\epsilon_2, 2+0.5\epsilon_1 + 0.5\epsilon_3)
\end{equation}
Assume that the input space is uniformly distributed and we choose to have two Gaussians($K = 2$) in our Gaussian mixture model approximation. To simplify the iteration process, we assume that the expectation-maximization step converges at Gaussians $\mathcal{N}_1((\frac{11}{18}, \frac{29}{18}), (\frac{1}{81}, \frac{1}{81}))$ and $\mathcal{N}_2((\frac{25}{18}, \frac{43}{18}), (\frac{1}{81}, \frac{1}{81}))$ with $w_1 = \frac{1}{2}, w_2 = \frac{1}{2}$. That is, in this Gaussian mixture model approximation there are two abstract elements that form the abstract domain
\begin{equation}
    \begin{split}
    c^{\#}_1 = <(\frac{11}{18}, \frac{29}{18}), (\frac{1}{81}, \frac{1}{81}), \frac{1}{2}>\\
    c^{\#}_2 = <(\frac{25}{18}, \frac{43}{18}), (\frac{1}{81}, \frac{1}{81}), \frac{1}{2}> 
    \end{split}
\end{equation}
And we have the abstract transformer performing on two Gaussians to compute layer propagation in abstract domain such that $f^{\#}: C^{\#} \rightarrow D^{\#}$
\begin{equation}
    \begin{split}
    f^{\#}(c^{\#}_1) &= <\begin{pmatrix}
        2 & -1\\
        0 & 1
    \end{pmatrix}\begin{pmatrix}
        \frac{11}{18}\\
        \frac{29}{18}
    \end{pmatrix}, \begin{pmatrix}
        2 & -1\\
        0 & 1
    \end{pmatrix} \begin{pmatrix}
        \frac{1}{81}\\
        \frac{1}{81}
    \end{pmatrix}\begin{pmatrix}
        2 & -1\\
        0 & 1
    \end{pmatrix}^{\text{T}} , \frac{1}{2}>\\
    &=<(-\frac{7}{18}, \frac{29}{18}), (\frac{1}{81}, \frac{1}{81}), \frac{1}{2}>
    \end{split}
\end{equation}
\begin{equation}
    \begin{split}
    f^{\#}(c^{\#}_2) &= <\begin{pmatrix}
        2 & -1\\
        0 & 1
    \end{pmatrix}\begin{pmatrix}
        \frac{25}{18}\\
        \frac{43}{18}
    \end{pmatrix}, \begin{pmatrix}
        2 & -1\\
        0 & 1
    \end{pmatrix} \begin{pmatrix}
        \frac{1}{81}\\
        \frac{1}{81}
    \end{pmatrix}\begin{pmatrix}
        2 & -1\\
        0 & 1
    \end{pmatrix}^{\text{T}} , \frac{1}{2}>\\
    & = <(\frac{7}{18}, \frac{43}{18}), (\frac{1}{81}, \frac{1}{81}), \frac{1}{2}>
    \end{split}
\end{equation}
The corresponding probabilities of two abstract elements given that the probabilities of data points are $p_{x^{(1)}}, p_{x^{(2)}} ...$ are 
\begin{equation}
    p_{c^{\#}_1} = \sum_i r_{i1} p_{x^{(i)}}, \quad p_{c^{\#}_2} = \sum_i r_{i2} p_{x^{(i)}}
\end{equation}
where $r_{i1}, r_{i2}$ are computed from the expectation-maximization iteration process in the convergence state. Obviously, both tend to $1/2$ in this problem setting. These two corresponding probabilities are also corresponding probabilities of $f^{\#}(c^{\#}_1)$ and $f^{\#}(c^{\#}_2)$ respectively such that
\begin{equation}
    p_{c^{\#}_1} = p_{f^{\#}(c^{\#}_1)}, \quad p_{c^{\#}_2} = p_{f^{\#}(c^{\#}_2)}
\end{equation}

\section{Conclusions and Future Work}
In this paper, we introduced distribution approximation and clusters approximation as two novel abstract domains which can be chosen with respect to different program circumstances and different properties concerned. We presented different distributions can be selected in distribution approximation such as polynomials, radial basis functions, and Fourier transform. We also presented different clustering methods can be selected in clusters approximation such as K-means and Gaussian Mixture Model. Comparing to grids approximation, these two methods can be more widely used for different circumstances.\\
We believe the key contribution of this work is that these two novel approximation methods not only work for purpose of neural network analysis, but also can be a general idea as an expansion of probabilistic abstract interpretation domains to solve other problems. \\
In the future, we will explore applications of two approximations in real world problems that can be analyzed by the framework. More state-of-the-art architectures of neural networks are left to be explored by the two approximation methods.

\end{document}